\documentclass{article}
\usepackage[utf8]{inputenc}
\usepackage{lmodern}
\usepackage{fullpage}
\usepackage{amsmath,amssymb,nicefrac}
\usepackage{booktabs,makecell}
\usepackage[usenames,dvipsnames]{xcolor}
\usepackage{soul}
\usepackage{graphicx}
\usepackage{authblk}

\usepackage[inline]{enumitem}
\usepackage{hyperref}

\newcommand{\sstt}[1]{{\texttt{\scriptsize #1}}}
\usepackage{relsize}
\newcommand*{\defeq}{\stackrel{\mathsmaller{\mathsf{def}}}{=}}

\title{Federated Multilingual Models\\for Medical Transcript Analysis}
\author{Andre Manoel$^\dagger$}
\author{Mirian Hipolito Garcia$^\dagger$}
\author{Tal Baumel$^\dagger$}
\author{Shize Su}
\author{Jialei Chen}
\author{Dan Miller}
\author{Danny Karmon}
\author{Robert Sim\footnote{Corresponding author: \url{rsim@microsoft.com}. $^\dagger$Authors contributed equally.}}
\author{Dimitrios Dimitriadis}
\affil{Microsoft Corporation}

\begin{document}
\bibliographystyle{IEEEtran}
\maketitle

\abstract{
Federated Learning (FL) is a novel  machine learning approach that allows the model trainer to access more data samples, by training the model across multiple decentralized data sources, while data access constraints are in place. Such trained models can achieve significantly higher performance beyond what can be done when trained on a single data source. As part of FL's promises, none of the training data is ever transmitted to any central location, ensuring that sensitive data remains local and private. These characteristics make FL perfectly suited for large-scale applications in healthcare, where a variety of compliance constraints restrict how data may be handled, processed, and stored. Despite the apparent benefits of federated learning, the heterogeneity in the local data distributions pose significant challenges, and such challenges are even more pronounced in the case of multilingual data providers. In this paper we present a federated learning system for training a large-scale multi-lingual model suitable for fine-tuning on downstream tasks such as medical entity tagging. Our work represents one of the first such production-scale systems, capable of training across multiple highly heterogeneous data providers, and achieving levels of accuracy that could not be otherwise achieved by using central training with public data. Finally, we show that the global model performance can be further improved by a training step performed locally.
}

\section{Introduction}
Federated learning is an ML paradigm for training models on decentralized data found in segregated  sources (silos), as described in~\cite{McMahan+16, Kairouz+19}.  The models are trained locally for a number of steps,  and then  combined together on a central server, ultimately creating a global model that contains information from all data sources. A basic assumption for most of these FL methods is the data sources are independent and identically distributed (i.i.d.) or at least with ``overlapping'' distributions. As such, these algorithms are not designed to adequately handle multilingual data, that is non-i.i.d.\ by definition, with severely skewed local distributions. In such scenarios, the  data in each silo is expected to be locally homogeneous-- usually i.i.d.,  monolingual data-- but the overall global data distribution remains non-i.i.d., making it challenging to aggregate all the local models. Consider, for example, the task of combining models independently trained on English and German corpora. Furthermore, the volume of available data per language can also be severely skewed-- high-resource language data can be much more abundant, making the trained model heavily ``focused'' on the high-resource languages and failing to adequately model the low-resource ones.

``\emph{Natural Language Understanding}'' (NLU) refers to machine extraction of knowledge from unstructured human communications, mainly text-based sources. Although NLU tasks have  been focused mostly on documents written in a single language, the joint analysis of multilingual documents is attracting increasing attention. It is shown that NLU models, when trained on multilingual datasets, can extract complementary knowledge from such vastly different corpora, improving the overall model performance~\cite{HLZ20}. In this context, the goal of multilingual NLU is to create a single  model for all languages, exploiting any correlations and underlying relationships that span beyond the language barrier. Most existing work on multilingual NLU is focused on scenarios where the  data is  centrally stored. However, multilingual data found in real-life scenarios, especially in the space of healthcare, is most often distributed, e.g., as in~\cite{Wang+22}. The participating data providers possess text data that is segregated and stored locally, and the NLU objective is to collectively process the documents without sharing any of the raw data. 
In this paper we present one of the first  commercial FL applications training a single global model in a multi-lingual healthcare setting (supporting nine data-segregated languages). 

Statistical NLU systems have been designed by extracting features from corpora using statistical and machine learning algorithms and they have gradually replaced traditional rule-based systems because of their superiority in generalization and robustness. In  healthcare, NLU is most usually applied to process medical-related text, such as clinical notes and other related text data. Clinical notes come from all medical scenarios and mainly consist of unstructured text stored in ``\emph{electronic health record}'' (EHR) systems, including medical notes, diagnostic reports, electronic prescriptions, \emph{et cetera}. Other text data may derive from other healthcare scenarios, e.g., clinical trials protocols, medical publications, surveys in population screening and articles for evidence-based reference. Besides these notes, an emerging source of data is based on transcriptions of  patient-doctor communications combined with machine translations and user-facing conversational bots~\cite{BRSK20}. Research on  applications of NLU for smart healthcare have received intensive attention in recent years, with some systems reaching maturity (in terms of productization)~\cite{BTRN19} among others. 

As mentioned above, in federated learning, a shared global model is trained under the coordination of a central server while keeping the user data segregated on local silos (aka clients).  Federated Learning has been applied to problems in NLU since its inception,~\cite{Liu+21, Lin+21}, and in particular for language modeling tasks~\cite{Yang+18, Hard+18, StSi20}. However, the multilingual NLU setting appears far more challenging. In such case, FL-based optimization suffers from training instability, slower convergence rates and lack of fairness for the smaller clients. On the other hand,  since FL techniques can now provide theoretical warranties for convergence in the case of non-i.i.d. data distributions, as in~\cite{LHYWZ20}, the value of its application to multilingual tasks takes on greater interest, where privacy and legal constraints are also of concern. Most of the legal constraints are based on the data sovereignty principle -- each data provider maintains  ownership and control of their data and, as such, it's not possible for the data to be mixed. The multilingual setup in healthcare falls under this constraint where the data silos are held by different providers, and even located in different countries with differing regulatory regimes.  Herein, we present a production-ready system overcoming the geopolitical barriers using techniques from FL optimization. Data in each silo is in a different language, and cannot be shared due to being sensitive, to the laws in different countries etc.  

Concurrent with the growing interest in Federated Learning, NLU has rapidly shifted towards the use of ``\textit{foundation language models}'' (fLMs), for foundation models see~\cite{Bommasani+22} and extensions to fLMs in~\cite{DCLT19}, GPT-3~\cite{Brown+20}. These fLMs are used as a starting point for learning other downstream NLU tasks. Such training strategies have become the golden standard for most of the concurrent applications. Lately, multilingual versions of these  models have been also proposed and are often used along with few-shot and/or transfer learning techniques to increase performance for tasks where the available target-language training data is limited. This state-of-the-art setup exploits the strong few-shot learning capabilities of large transformer models and fLMs generally. The scenario herein presented is one occurrence of such federated learning approaches for fLMs. Other papers in this space focus on the performance gap between federated learning  and centralized training, evaluating on a wide variety of English NLU tasks~\cite{LiMi20, Lin+21}. On the contrary, we differ from such work by studying the federation of fLMs in a highly imbalanced and non-i.i.d. setup, with performance constraints across all participating languages.
  
In this work, we explore multilingual Federated Learning across 9 languages, each with various amounts of available training data, i.e. Table~\ref{table:public}, while leveraging a pretrained foundation model as the initial/seed model. Our results show that fine-tuning such pretrained models with FL can perform similarly to the standard centralized methods (in the case when  no data accessibility constraints are in place), despite having completely non-i.i.d. data distributions among the participating silos (each with monolingual data). We show that training the fLMs this way provides an effective and generalizable way for processing multilingual data all while benefiting from the accessibility features of FL at little or no cost to the final/downstream task performance. As part of the proposed solution, accessing and sampling the individual silos based on the available resources per language, can ensure a more fair knowledge representation.

In addition to the sampling strategy, we investigate the merits of personalization in the overall model performance. We show that under-resourced languages can benefit either from simple fine-tuning or from model interpolation according to the deployed algorithm, more details in~\cite{DKMM20}.

The contributions of the presented system are as follows:
\begin{enumerate}
\item We present one of the first production-ready FL-based systems for medical-related NLU where the models are trained on real-life data. 
\item An end-to-end FL system for training base multi-language models for medical texts understanding, when the participating clients (in different languages) have different amounts of training data. Herein, we  present different sampling strategies for the proposed system.
\item Investigate when personalization in the form of model interpolation  can benefit performance, especially for under-represented languages.
\item Experimental validation of the proposed approach: we present comprehensive experimental results supporting the proposed design and algorithmic decisions.
\end{enumerate}

In summary, this work presents the federation of an NLU foundation model trained on multilingual data for medical text analytics as the down-stream task. The challenges for such  systems mostly lie in the extreme non-iidness of the multilingual data, the skewed training data distributions in terms of volume and the data constraints driven by the geopolitical legal framework. The deployed system federates a single multilingual model without mixing data from different silos and/or locales, all while improving performance for all locales in tandem, Table~\ref{table:perplexity}. The data sources and algorithms implemented in this system are detailed in Section~\ref{sec:methods}, including the design decisions for client sampling, optimization, etc. Subsequently, the experimental findings are presented in Section~\ref{sec:results}. We  focus mainly on the performance of the trained models on the downstream task, i.e. ``\emph{Text Analytics for Healthcare}'' (TA4H), as well as the impact of local personalization, i.e. fine-tuning the global model on each language. These results are broken down for each of the available languages and discussed in Section~\ref{sec:discussion}, and finally a discussion about the legal framework governing Healthcare is presented. 

\section{Methods}
\label{sec:methods}
\subsection{Task and Model Description}

Pretrained transformer-based language models~\cite{vaswani2017attention, devlin2018bert} achieve state-of-the-art performance on a variety of NLU tasks \cite{wang2018glue, wang2019superglue}, especially when pretrained on domains similar to the target task~\cite{gururangan2020don}. Such models also show great promise achieving zero-shot classification~\cite{liang2020xglue} when trained across multiple languages~\cite{conneau2019unsupervised, jiang2021xlm}.

The proposed approach to achieve state-of-the-art performance on multilingual medical domain NLU tasks is to further fine-tune an already pretrained model, i.e. an XLM-K~\cite{jiang2021xlm} model, on the medical domain, with a specific focus on multilingual clinical notes. At the time of writing, we have only been able to find such models in English~\cite{beltagy2019scibert, lee2020biobert, pubmedbert}. In fact, one of the key problems we faced when trying to build this model is the availability of public medical text in other languages, especially clinical notes. The training task we chose for fine-tuning the model is ``Masked LM'', where some of the words in the sentences are randomly masked and the model needs to accurately predict them. As such, no additional labels are required for the training phase. 

Once this new model is obtained, we can fine-tune it to perform other downstream tasks.
This multilingual model is also fine-tuned on the medical domain and as such, it is the first of its kind. 
Our experiments were focused on ``\textit{named entity recognition}'' (NER), i.e., identifying whether there are words or expressions on the notes that belong to one of a few categories like symptoms, diseases, medical codes, and so on.

\subsection{Data Sources}
\label{sec:data}
All the data used in this paper except proprietary clinical notes and proprietary non-English annotated texts is publicly available; review Table \ref{table:public} for un-annotated texts used for training the language models and Table \ref{table:ner_public} for NER datasets. All texts were tokenized using NLTK\cite{bird2009natural} and custom scraping tools (wikidata query tools\footnote{\url{https://query.wikidata.org/}}, selnium\footnote{\url{https://www.selenium.dev/}}, Beautiful Soup\footnote{\url{https://www.crummy.com/software/BeautifulSoup/bs4/doc/}}, E-utilities\footnote{\url{https://www.ncbi.nlm.nih.gov/books/NBK25500/}}) were developed to obtain the un-annotated texts.

\begin{table}[ht!]
\centering
\begin{tabular}{crrrl}
    \toprule
    silo id & language & \# train samples & \# test samples & sources \\
    \midrule
    1 & Italian & 412,437 & 26,192 &  \makecell[l]{e3c\cite{magnini2020e3c}, EMA\cite{gotzsche2011opening},\\ Wikipedia clinical articles\cite{wikipedia2004wikipedia}} \\ \hline
    2 & French & 477,323 & 31,167 &  \makecell[l]{ PMC\cite{roberts2001pubmed}, QUAERO\cite{neveol2014quaero}, EMA, \\ Wikipedia clinical articles}\\ \hline
    3 & Spanish & 502,479 & 42,225 &  \makecell[l]{PMC, EMA, SciELO\cite{packer2009scielo},\\ CODEIESP\cite{miranda2020codiesp}, e3c,\\
    Wikipedia clinical articles}\\ \hline
    4 & Portuguese & 400,175 & 26,192 &  \makecell[l]{PMC, EMA, \\ Wikipedia clinical articles} \\ \hline
    5 & English & 132,593,658 & 4,400,378 & \makecell[l]{ClinicalTrials.gov\cite{zarin2011clinicaltrials}, \\ MIMIC-III\cite{johnson2016mimic}, MIMIC CXR\cite{johnson2019mimic}, \\ i2b2\cite{uzuner2006i2b2}, Wikipedia clinical articles,\\ Propriety Clinical Notes} \\ \hline
    6 & German & 575,559 & 26,192 &  \makecell[l]{PMC, OPUS\cite{tiedemann2012parallel}, \\ Wikipedia clinical articles} \\ \hline
    7 & Arabic & 83,963 & 26,192 &   \makecell[l]{Wikipedia clinical articles}\\ \hline
    8 & Hebrew & 34,104 & 26,192 &   \makecell[l]{IMA\cite{ima}, MyTrails\cite{mytrail}, \\ Wikipedia clinical articles}\\ \hline
    9 & Russian & 100,762 & 26,192 &  \makecell[l]{PMC, Wikipedia clinical articles} \\ \hline\hline
    & & \textbf{135,280,160} & \textbf{4,630,922} & \\
    \bottomrule
\end{tabular}
\caption{Public data used to validate the FL model. Notice that data is severely unbalanced across silos: the one containing English data has many more samples than all of the remaining. The procedure for transforming texts collected from sources into lists of sentences is explained in Section \ref{sec:data}.}
\label{table:public}
\end{table}

\subsection{Baseline Model}

The central learning (CL) baseline is obtained with PyTorch based on a conventional optimization process for pre-training using the data collected from different sources. All hyperparameters are tuned according to previous experiments with similar models and datasets, and are detailed in Table~\ref{table:hyperparams}. The results shown in Table~\ref{table:perplexity} and Figure~\ref{fig:compar} refer to the final model obtained, mentioned as CL model/results.

Roughly 10.4\% of the training data is used during the process, since the volume of available data is too large, and using all of it  was deemed prohibitive. Batches are created by sampling uniformly at random from examples from all languages, and are not stratified per language. 
Evaluation is performed every 2048 batches of training, using 65,536 samples picked at random from the test set. These validation perplexities are reported in Figure~\ref{fig:compar}-- on the contrary, the perplexities reported in Table~\ref{table:perplexity} are computed over the full test set.

In parallel, we trained a single model per silo (without either FL or mixing data from other languages) with the corresponding results shown in Table~\ref{table:perplexity}. The exact same hyperparameters and amount of data are herein used, i.e., 10.4\% of the data available on the silo, etc. 

\begin{table}[ht]
\centering
\begin{tabular}{ccc}
    \toprule
     & CL/per-silo & FL \\
    \midrule
      \# samples seen & 
      \makecell{ \scriptsize{Per batch:} \\[-0.5ex] \small{2048} } & 
      \makecell{ \scriptsize{Per round on silo $i$:} \\[-0.5ex] \small{$\max(500, 0.8 \cdot 10^{-4} N_i)$} \\[0.1ex] \scriptsize{Per batch:} \\[-0.5ex] \small{2048} } \\ \midrule
      optimizer & 
      \makecell{ \sstt{AdamW} \\[-1ex] \sstt{lr ($\gamma_0$) = 3e-5} \\[-1ex] \sstt{eps = 1e-6} \\[-1ex] \sstt{weight\_decay = 0.01} } & 
      \makecell{ {\scriptsize Client/Silo:} \\[-1ex] \sstt{SGD} \\[-1ex] \sstt{lr = 1e-4} \\[-0.5ex] {\scriptsize Server:} \\[-1ex] \sstt{Adam} \\[-1ex] \sstt{lr ($\gamma_0^s$) = 3e-4} } \\ \midrule
      LR scheduler &
      - &
      \makecell{ \scriptsize{At round $r$:} \\[-1ex] \sstt{$\gamma^s = (1 - 10^{-3}r) \, \gamma_0^s$} } \\
    \bottomrule
\end{tabular}
\label{table:hyperparams}
\caption{Hyper-parameters used for pre-training the model.}
\end{table}

Next, we provide more details about the training of the federated model.

\subsection{Federated Learning}
\label{sec:fl}

Available medical data is not fully exploited by researchers and medical institutions given the constraints on transmitting sensitive information from private silos to a centralized location. Federated Learning provides a collaborative learning environment with privacy guarantees under the coordination of a central server.
Herein, the proposed  production-scale system can train the models across hundreds of silos without sharing raw data, allowing  partners  across different storage subscriptions of  training a single model.

For performing the federated training, we use a hierarchical optimization approach~\cite{Dimitriadis+21}, where a persistent optimizer stays on the server,
$f^s(\cdot)$; and intermittent stateless ones $f^c_i(\cdot)$, one per  silo $i$,  are re-instantiated at each round and silo. This approach has been shown to improve the convergence rate, allowing better control of the learning process. Specifically, the global model at round $r$, $\theta^{(r)}$ is communicated to the participating silos $i$ and trained on their data, with the the successive local gradients 
$g^{(r, b)}_i$, $b = 1, \dots, B$. The silo-side optimizers then update the local
weights $\theta_i^{(r, b + 1)} = f^c_i (\theta_i^{(r, b)},\ g^{(b)}_i)$, with $\theta_i^{(r, 0)} \equiv 
\theta^{(r)}$. Once the maximum number of local batches or the end of local data is reached,  the silo estimates a local ``pseudo-gradient'' $g_i^{(r)}$,  
\begin{equation}
    g_i^{(r)} \defeq  \theta^{(r)} - \theta_i^{(r, B)},
\end{equation}
and transmits it back to the server. Once all silos have transmitted their local gradient, the updated global model is given by 
\begin{equation}
    \theta^{(r + 1)} = f^s \big(\theta^{(r)},\ \sum_i w_i g_i^{(r)})\big).
\end{equation}
The weights $w_i$ that can be set in different ways~\cite{Dimitriadis+21} -- in most experiments, we have used $w_i = N_i / \sum_i N_i$, as in FedAvg~\cite{McMahan+16}. A number of hyperparameter values have been investigated, and the ones in Table~\ref{table:hyperparams} have been picked based on the best perplexity results during validation\footnote{A side-comment here is that the final results are quite sensitive to the batch sizes.}.

Only a fixed amount of $\max(500, 0.8 \cdot 10^{-4} N_i)$ samples is used per iteration, regardless of the total number of available samples $N_i$ per silo $i$. These local training samples are picked uniformly at random, \emph{with} replacement. This is a deviation  from the sampling approach in the CL scenario, where samples are seen once per epoch.
The effective batch size for our model is based on the fact that only 500 samples per silo are used at each given round. The only exception is the silo containing the English data being ``throttled'', where $10.6k$ samples are processed in the form of 6 batches per round.
As mentioned, the number of samples per silo/language at each iteration can significantly affect the final results; for example, a different setup using $\max(100, 10^{-4} N_i)$ samples per language resulted in the English language being over-represented, and the model that was over-fitted to English and under-fitted to the remaining languages.

Validation is performed every 5 FL rounds, with 10\% of the total amount of test samples being picked at random. These validation results are shown in Figure~\ref{fig:compar}; the results on Table~\ref{table:perplexity} are based on the final model and reported on the entire test set.

\subsection{Personalization}
\label{subsec:methods_personal}

The convergence of most  Federated Learning optimization algorithms is  theoretically proven when the client data distributions are iid. However, scenarios, such as the multilingual NLU ones, where the data distributions are non-iid, are far more challenging. One of the different approaches for addressing this issue is with convex interpolation between the global $\theta^{(r)}$ and the local models $\theta^{(r,B)}_i$,~\cite{DKMM20}. The resulting model $\theta^{(r)}_{int}$ after interpolation is given by    
\begin{equation}
    \theta^{(r)}_{int} = \alpha_i \cdot \theta^{(r, B)}_i+(1-\alpha_i) \cdot \theta^{(r)}
\end{equation}
and the interpolation weights $\alpha_i$ for each client $i$ are estimated as described in~\cite{DKMM20}. 

We have tried different strategies for training the local models depending on the initial checkpoint of the local model. We have found that starting the local model training later in the FL process allows for improved generalization of the models. Otherwise, the local models are quickly overfitting, degrading the overall performance. More details are discussed in Section~\ref{subsec:results_personal}. 

\subsection{Engineering System Architecture} %
\label{subsec:engineering}

The proposed solution is based on Azure Arc-enabled Kubernetes clusters\footnote{More details about Kubernetes can be found in \href{https://github.com/Azure/AML-Kubernetes}{\underline{AML-Kubernetes}} and the Azure Machine Learning (AML) platform, in \href{https://techcommunity.microsoft.com/t5/ai-machine-learning-blog/realizing-machine-learning-anywhere-with-azure-kubernetes/ba-p/3475298}{\underline{AML Platform}}. }, enabling large-scale FL applications on the Cloud.
The developed FL system offers templates for the target task and a public API\footnote{More details in \href{https://shrike-docs.com/pipeline/federated-learning-doc/}{\underline{FL API documentation}}.} allowing easy-to-deploy in AML real-life FL tasks \footnote{More details and FL examples can be found in the public repo \href{https://github.com/Azure-Samples/azure-ml-federated-learning}{\underline{AML FL github repo}}}.

\begin{figure}[ht]
    \centering
    \includegraphics[width=0.5\textwidth]{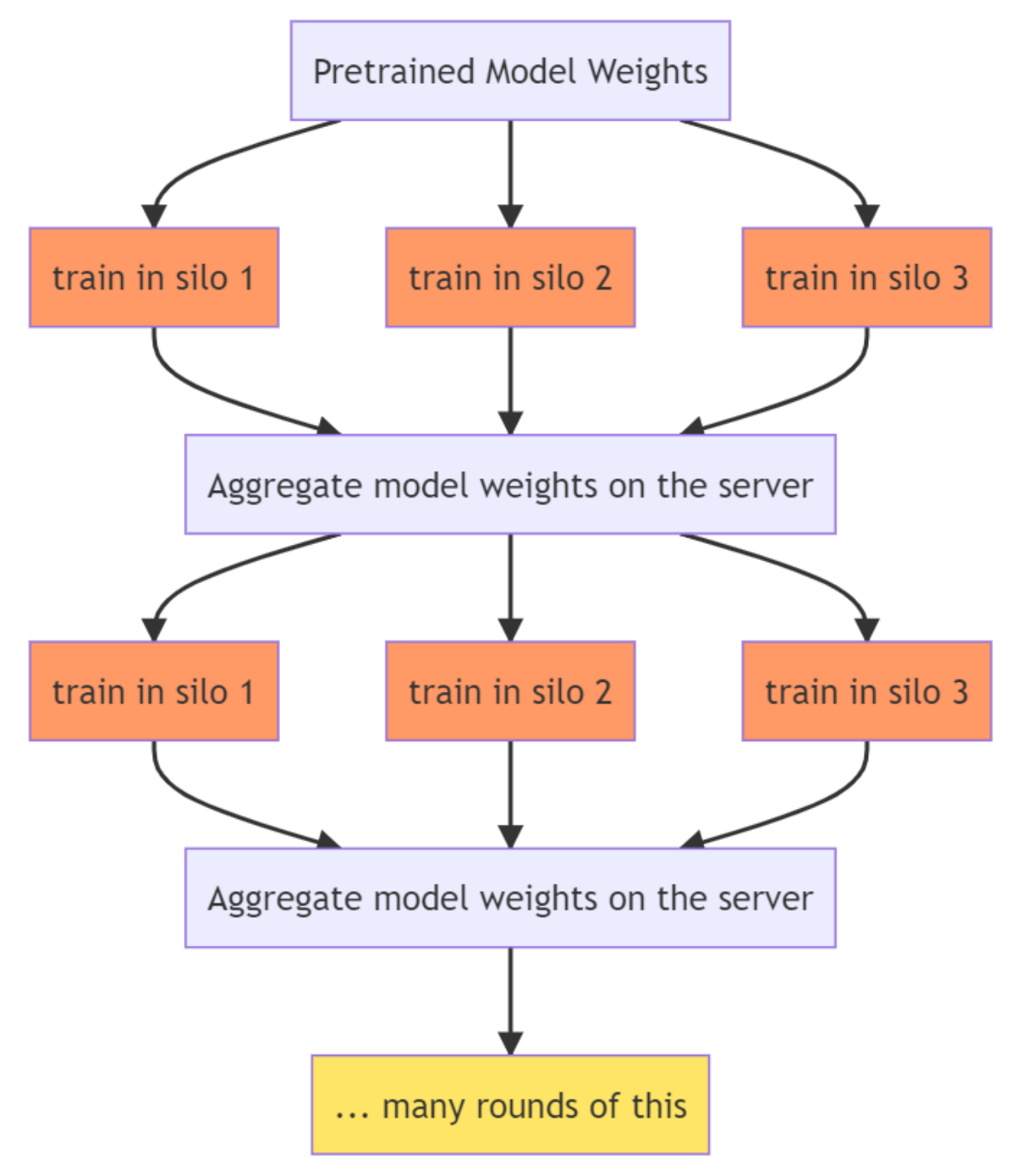}
    \caption{Cross-silo federated learning flowchart (basic version without using secure aggregation).}
    \label{fig:flowchart}
\end{figure}

Following the FL principles, the local data never leave the customer tenant, and are only processed on their own Arc-enabled Kubernetes compute cluster. Each silo trains iteratively a local version of the global model using its own data on their own compute environment. At each iteration, a silo might either use the silo's full data for the model training, or a randomly sampled subset (and different subsets across iterations) of the silo's data. Once these silos finish processing their data (or reach a maximum number of processed batches), the locally adapted model weights are transmitted back to the central orchestrator/server. On the server, the model weights from all silos are then aggregated appropriately and update the global model of the previous iteration, before moving to the next training iteration.
The flowchart is shown in Figure~\ref{fig:flowchart}.

 In more detail, the FL task is set up according to the published recipes~\footnote{\url{https://github.com/Azure-Samples/azure-ml-federated-learning}}, 
 with the required resources: 
 \begin{enumerate}
\item  the central orchestrator in an AML workspace, 
\item the compute environment for this central orchestrator, 
\item the individual data silos with azure arc-enabled kubernetes compute environments, 
\item the azure data factory compute for transferring intermediate model weights between silos, and 
\item central orchestrating compute server
\end{enumerate}. 
Next, the users can create the FL pipeline by following the public API documentation, i.e., simply provide a federated learning configuration file (specifying parameters like `max\_iterations', compute targets and input dataset names for each silo, etc.), and create the appropriate component functions, as shown in Figure~\ref{fig:flowchart}, e.g., `train' for local model training at each silo, `midprocess' for the model weight aggregation at server, etc. The available FL API also provides  visualization tools for the AML workspaces.

Finally, in a more recent version of the FL API  another layer of security has been added by applying multi-party compute secure model, when additional privacy and security is necessary/desired. This method uses cryptographically generated masks, based on ``\emph{Secure Multi-party  Computation}'' (MPC)~\cite{CFGN96}, before transferring the weights from the silos to the server, ensuring the server cannot link the communicated model weights to a particular client. The added masks are designed to cancel each other during the aggregation step, and as such the aggregated model will be accurate\footnote{More details about the process can be found in the public  API documentation \href{https://shrike-docs.com/pipeline/federated-learning-doc/}{public FL API documentation} and the public federated learning recipes (https://github.com/Azure-Samples/azure-ml-federated-learning)to implement such a large scale practical federated learning pipeline for experiments on the cloud using Azure Machine Learning.}. 

\section{Results}
\label{sec:results}
\subsection{Validation on Public Data}

We first use publicly available data to demonstrate the capacity for a multi-lingual fLM to pretrain in a federated setting with results competitive with central training.  Starting with a base snapshot of XLM-K~\cite{jiang2021xlm}, we federate the model across nine silos, each of which contains medical text in a distinct language as shown in Table~\ref{table:public}. The data inside each silo comes from different public sources, which were compiled into proper datasets following the procedure described in Section \ref{sec:data}.

\begin{table}[t]
\centering
\begin{tabular}{c|rr|rrrrrrrrr|r}
    \toprule
    lang. & FL & CL & It & Fr & Es & Pt & En & De & Ar & He & Ru & Base \\
    \midrule
    It & \textbf{5.63} & 5.80          & {\color{BrickRed} 6.32}  & 8.12  & 10.18 & 8.51  & 22.12 & 8.11  & 9.22  & 10.76 & 9.25  & 16.10 \\
    Fr & 4.97          & \textbf{4.94} & 7.16  & {\color{BrickRed} 5.53}  & 7.62  & 7.36  & 18.09 & 6.98  & 8.01  & 9.35  & 7.72  & 13.13 \\
    Es & \textbf{5.65} & 5.81          & 9.01  & 7.96  & {\color{BrickRed} 6.45}  & 7.74  & 17.93 & 8.11  & 9.17  & 11.40 & 9.11  & 17.28 \\
    Pt & \textbf{5.85} & 6.01          & 9.36  & 8.91  & 9.40  & {\color{BrickRed} 6.91}  & 22.68 & 8.98  & 10.23 & 12.46 & 10.31 & 18.65 \\
    En & \textbf{3.41} & 3.51          & 15.60 & 14.65 & 13.84 & 14.57 & {\color{BrickRed} 3.50}  & 13.96 & 15.88 & 20.71 & 15.54 & 32.83 \\ 
    De & 7.10          & \textbf{7.04} & 10.38 & 10.07 & 10.24 & 10.47 & 28.12 & {\color{BrickRed} 7.74}  & 11.57 & 14.60 & 11.42 & 21.98 \\
    Ar & \textbf{8.36} & 10.48         & 12.07 & 12.30 & 12.20 & 12.66 & 33.79 & 11.25 & {\color{BrickRed} 10.81} & 13.78 & 12.22 & 19.82 \\
    He & \textbf{6.98} & 9.44          & 10.44 & 10.14 & 10.36 & 11.18 & 32.08 & 9.73  & 10.80 & {\color{BrickRed} 11.43} & 10.65 & 19.30 \\
    Ru & \textbf{6.07} & 7.17          & 7.95  & 7.66  & 7.59  & 8.12  & 25.95 & 7.56  & 8.65  & 10.39 & {\color{BrickRed} 7.34}  & 14.77 \\
    \bottomrule
\end{tabular}
\caption{Perplexity per-language obtained on the test data, using multiple different models: the one pretrained with federated learning (FL); another pretrained with centralized learning (CL), where all data is pooled together; and others obtained by using only data from specific silos. \emph{Base} indicates perplexity evaluated on the base model snapshot. Note that the best result is always better than the one obtained using only data in that language (in red), showing that other languages indeed contribute to the performance of the model.}
\label{table:perplexity}
\end{table}

As usual in FL, the pretraining procedure consisted in a sequence of iterations where, at each iteration, every silo would produce a new model and send it to the server, which then combines all models it received. Procedures for updating the models and for combining them are detailed in Section \ref{sec:fl}. The system infrastructure for the silos, server and the coordination, are described in Section \ref{subsec:engineering}.

\begin{figure}[!b]
    \centering
    \includegraphics{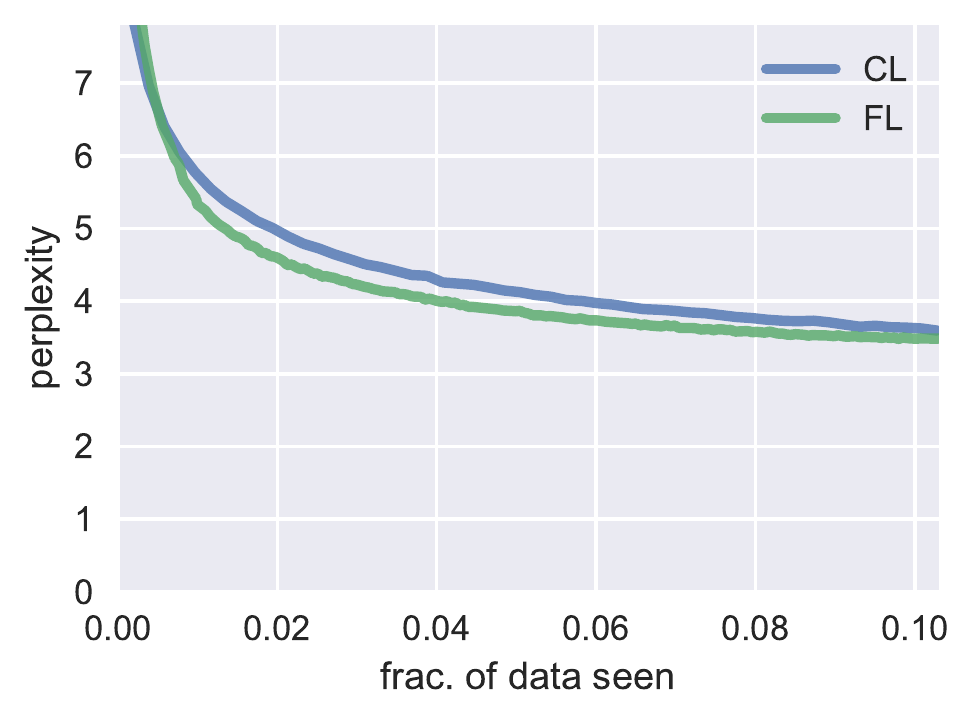}
    \caption{Convergence of FL vs CL in terms of the overall perplexity over the test set. After approximately 10\% of the data has been seen, the two models report similar perplexities, of 3.49 (FL) and 3.59 (FL). Small differences can be attributed to the batch stratification that happens in FL, as well as the choice of hyper-parameters, which were independently optimized for each model.}
    \label{fig:compar}
\end{figure}

Per-language model quality was assessed using the perplexity on test data, which for simplicity was made available to the server. The FL performance was compared to four different baselines: one produced using central learning (CL), with the data from all silos pooled together; and other three using only data from specifics silos, English (En), German (De), and Spanish (Es), the three languages with most samples available.
In addition, we report perplexity results for the zero-shot task (i.e.\ performance using the base model snapshot). 
Importantly, all models were pretrained using roughly the same amount of data, corresponding to approximately 10\% of the total number of samples. The hyper-parameters for FL and CL were independently adjusted, and the per-silo models used the same parameters as CL.

The results in Table~\ref{table:perplexity} show FL performance competitive with and sometimes outperforming central training. We also observe that English, German, and Spanish all benefit from training on all languages, rather than training per-language models. Figure~\ref{fig:compar} illustrates the convergence of the federated model compared to central training.

Finally, we used the FL and CL snapshots to evaluate performance on downstream tasks. Specifically, we have fine-tuned two models for performing named entity recognition (NER), each starting from one of the snapshots. This procedure was repeated for three different public datasets: BC5CDR\cite{BCDR}, NCBI-disease\cite{dougan2014ncbi}, and i2b2\_2009\cite{patrick2010high}. As seen in Table \ref{table:ner_public}, both give similar results in terms of achieved micro, macro and weighted f1-scores.

Together, Tables \ref{table:perplexity} and \ref{table:ner_public} and Figure \ref{fig:compar} show that our federated learning setup can produce comparable, if not superior, models, without sharing raw data during the training process. This result has already been observed in multiple works across the literature, although most consist of simulations that cannot be easily generalized to real-world settings. Within our setup, this is just a matter of replacing the silos containing public data with real-world data silos, since they are already on isolated cloud environments. This is what we have done next.

\begin{table}[ht]
\centering
\begin{tabular}{cccc}
\toprule
dataset & FL & CL \\
\midrule
BC5CDR & 0.8913/0.8917/0.8915 & \textbf{0.8917}/\textbf{0.8920}/\textbf{0.8919} \\
NCBI-disease & 0.9287 & \textbf{0.9370} \\
i2b2\_2009 & \textbf{0.9184}/\textbf{0.8476}/\textbf{0.9170} & 0.9174/0.8446/0.9155 \\
\bottomrule
\end{tabular}
\caption{Best f1-score obtained performing NER on three different datasets. Different numbers give respectively micro, macro and weighted averages over entities (for  NCBI-disease a single number is presented, since there is a single entity).}
\label{table:ner_public}
\end{table}

\subsection{Deployment and Validation on Real-World Data}
The federated approach described in this paper was deployed in a real production environment  to train a multi-lingual fLM with Hebrew support, considering
the lack of publicly available clinical text in this language. Hebrew clinical notes were de-identified using HebSafeHarbor\footnote{HebSafeHarbor code \url{https://github.com/8400TheHealthNetwork/HebSafeHarbor}} and used to tune XLM-K to the medical domain, using 2.5gb of proprietary Hebrew clinical notes in one silo, and English public clinical text in a separate silo.

Once pretraining has been performed using federated learning, the resulting model was attached to different classification heads to fine-tune on proprietary data, for solving 3 tasks: \textit{"named entity recognition"} (NER), \textit{"assertion detection"} (AD) and \textit{"relation extraction"} (RE). This dataset consisted on annotated clinical notes in both English and Hebrew. Note that the de-identified clinical notes could not be accessed in a centralized training setting.

The average f1-score obtained in 2 of these 3 tasks is reported in Table \ref{table:hebrew_propitary_tasks}. The AD task has a highly unbalanced label set and we found the classification scores to be uninformative, so they are not reported here.  Note that for the reported tasks the FL model provided considerably higher accuracy in all cases, when compared to the previous model which used public data only.


\begin{table}[ht]
\centering
\begin{tabular}{cccc}
\toprule
Task & CL Public Datasets  &  FL De-identified Hebrew \\
\midrule
NER (strict/type) & 0.5415/0.6778 & \textbf{0.6628/0.8118} \\
Relation Extraction & 0.8144 & \textbf{0.8178} \\
\bottomrule
\end{tabular}
\caption{Comparison of f1-score obtained in downstream tasks with proprietary data, using either a language model trained on public clinical text in a centralized setting, or a language model trained on Hebrew/English silos in a federated setting. For the NER task, scores were computed in two different ways, requiring either exact entity boundaries to match (\emph{strict}) or only the entity types (\emph{type})\cite{segura2013semeval}.}
\label{table:hebrew_propitary_tasks}
\end{table}

\subsection{Personalization}
\label{subsec:results_personal}

Silos can leverage their local data to obtain models that are better adjusted to the local data distribution. In Table \ref{table:personalization}, we introduce a \emph{personalized} model, which starts from the global FL model at a given iteration, and only uses local data from that iteration onward. As shown in the table, the accuracy provided by this personalized model in the task of predicting masked tokens is often better than that of the FL model, especially for non-English silos. Moreover, if we interpolate between these two models, following the procedure described in Section \ref{subsec:methods_personal}, even better results can be obtained---the last row of the table shows results for such \emph{interpolated} model.

\begin{table}[ht]
\centering
\begin{tabular}{crrrrrrr}
    \toprule
    & English & French & Spanish & German & Russian & Portuguese & Italian \\
    \midrule
    Federated & \textbf{0.6976} & 0.6404 & 0.6183 & 0.5950 & 0.5876 & 0.6145 & 0.6194 \\
    Personalized & 0.6945 & 0.6491 & \textbf{0.6286} & 0.6007 & \textbf{0.6054} & 0.6213 & \textbf{0.6288} \\
    Interpolated & 0.6970 & \textbf{0.6510} & 0.6279 & \textbf{0.6024} & 0.6043 & \textbf{0.6220} & 0.6279 \\
    \bottomrule
\end{tabular}
\caption{Test accuracy obtained in predicting masked tokens, using different models. Personalized, silo-specific models typically provided a better performance than the global FL one; moreover, interpolating these two models can improve the performance even further.}
\label{table:personalization}
\end{table}

The experiment above was run as a proof of concept using the FLUTE simulation platform \cite{dimitriadis2022flute}, with the same public data as in previous experiments. After confirming that personalization and interpolation can indeed improve the performance of the global FL model, we ran additional experiments in the production platform on AML. Specifically, we trained a personalized model starting from the final FL model, using only German data; this, already, provided better performance when evaluating on the German test set. Moreover, interpolating between the global FL model and this personalized model provided even better results than using just one or the other, with the interpolation factor $\alpha$ set to 0.9.

\begin{figure}[ht]
    \centering
    \includegraphics[width=0.6\textwidth]{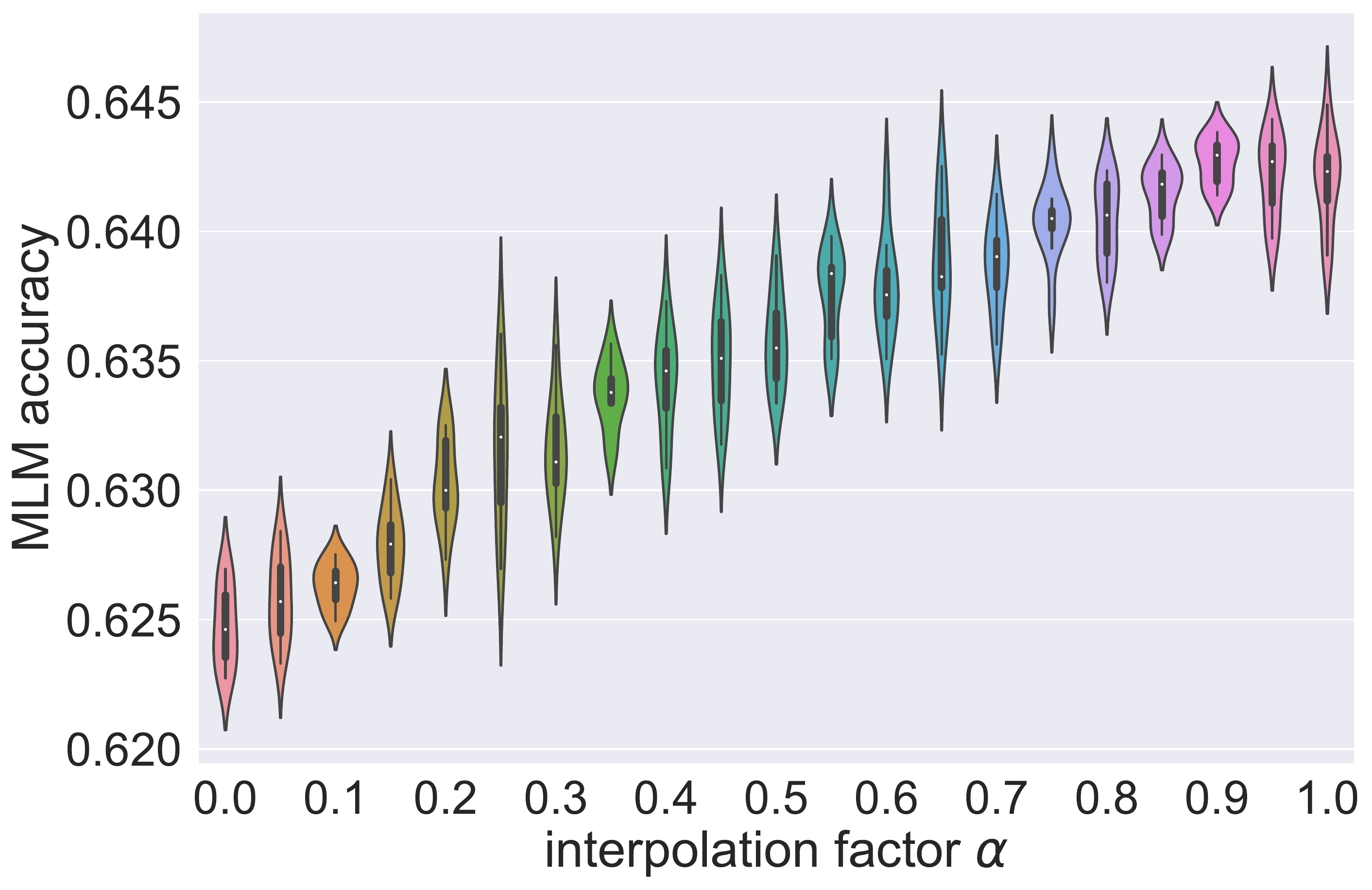}
    \caption{Test accuracy obtained on a masked language modeling (MLM) task, by interpolating between the global FL model ($\alpha$ = 0) and a personalized model ($\alpha$ = 1) trained post-hoc, starting from the global FL model and using only German data. For each of 21 values of $\alpha$ equally partitioning the range $[0,1]$, an evaluation over the whole German test set was performed. While the personalized model already has better performance than the global FL model on the German test data, interpolating can provide even better results, in this case with an $\alpha$ value of 0.9. See Section \ref{subsec:methods_personal} for more details on the interpolation procedure.}
    \label{fig:personalization}
\end{figure}

\section{Discussion}
\label{sec:discussion}
The current proposed approach navigates through two different sets of constraints and considerations: one set can be attributed to the legal and privacy framework and the Cloud business reality, and the second one attributed to the algorithmic/task-related challenges.

\subsection{Policy and Business Constraints}

\paragraph{Policy Challenges}
Today many parts of the world require technology companies to treat user data, which is usually generated and stored in data silos (e.g., service providers’ data centers), according to user-privacy laws. Examples of such  laws include the ``\textit{European Union General Data Protection Regulation}'' (GDPR)\footnote{\url{https://gdpr-info.eu/}}, the ``\textit{California Consumer Privacy Act}'' (CCPA)\footnote{\url{https://www.oag.ca.gov/privacy/ccpa}}, and the ``\textit{Health Insurance Portability and Accountability Act}'' (HIPAA)\footnote{\url{https://www.cdc.gov/phlp/publications/topic/hipaa.html}}. 
Such legal constraints pose a challenge to traditional centralized ML approaches, where all data is usually stored at a single, central location, typically a Cloud data center. 
Centralized data  management and processing can impact transparency and data provenance tracking, which in turn could lead to the lack of trust from the end-users, as well as increased difficulty in compliance with the data governance laws like the GDPR and/or HIPAA.  As a response to these challenges, Federated Learning is an evolving technology that is well positioned to address such policy challenges appropriately.

\paragraph{Federated Learning on Cloud ML Platforms}

Besides these legal constraints, more practical issues with centralized training are the need for specialized computing resources,  the fact that large-scale data collection and processing held on a single server can be seen as a single point-of-failure and  a non-negligible  risk of data breaches. Such hardware requirements drive increased costs from the service provider for training and maintaining the proper infrastructure.

Regardless, a number of incumbent cloud-based ML training platforms has emerged, including Azure ML.  These platforms, while highly capable for processing very large data volumes and machine learning tasks, are characterized by their dependence on centralized data for training. On the other hand, a few platforms (e.g., Flower\footnote{\url{https://flower.dev/}} and sherpa.ai\footnote{\url{https://www.sherpa.ai/}}, OpenMined\footnote{\url{https://www.openmined.org/}}) have been developed to enable data collaboration across security boundaries.  However, these solutions focus more on smaller-scale user (data scientist) interaction solutions without built-in integration with cloud providers. There are a few large scale federated learning platforms, some not open-sourced e.g., from Google, Apple and Meta, and open-sourced Microsoft Research, FLUTE\footnote{\url{https://github.com/microsoft/msrflute}}, as~\cite{dimitriadis2022flute},
for federated learning using consumer, mobile, advertising data, rather than cloud enterprise data. Our focus has been on building and deploying large-scale cross-silo federated learning on Cloud processing enterprise data across dozens or hundreds of data providers. Our implementation, which is based on bespoke public APIs for cross-silo job orchestration in AML\footnote{\url{https://github.com/Azure/AML-Kubernetes}}, is one of the first ones to address how existing ML platforms can be repurposed for cross-silo federated learning at scale.

\subsection{Algorithmic Challenges}

\paragraph{Cross-Silo FL with Unbalanced Data} 
Our experiments show that FL-based model training, with no or very limited impact in performance (when compared with the centralized training)\footnote{Centralized training can provided an upper bound in performance but it's infeasible in real-life scenarios where data accessing constraints are in place.} is possible even in settings where the local datasets are highly heterogeneous. 
While similar tasks have been extensively explored in the literature, the current work presents some innovative design decisions, e.g., the use of multilingual text data combined with a multilingual fLM in a healthcare context, a sampling strategy for the clients, etc. The proposed deployment is supported by the Azure ML platform as a straightforward extension. As detailed in Section~\ref{subsec:future},  some algorithmic challenges concerning the stability and robustness of the presented system are still pending. For example, we have noticed that the amount of data used at each silo on any given iteration can widely affect the final result, due to model under- or over-fitting across the different languages. Handling such instability requires additional innovation as part of the FL optimization. 

\paragraph{Personalization}
Finally, we investigate how personalization can improve performance in the presence of extreme statistical heterogeneity, as such in the case of multilingual NLU model training. In addition to the statistical heterogeneity of the local/silo data distributions, the investigated FL setup presents additional heterogeneity due to the training dataset imbalance, e.g., the number of English training examples is at least 2 orders of magnitude more than the rest of the multilingual data. As such, this imbalance of training examples skews the global models significantly, as shown in Table~\ref{table:personalization}. Tackling these  heterogeneity challenges, a straightforward solution is to maintain multiple  models for the different local distributions, e.g., as proposed in ``Clustered FL''~\cite{SMS20}. Another strategy is for FL-based personalization~\cite{TYCY22}, where a different personalized model is generated for each client by leveraging both global and local information. Herein, we  capture the  differences in the local distributions by interpolating two models, the global and a local one (one local model per language), as described in~\cite{DKMM20}. 

It is shown that  standard FL algorithms fail to ensure fairness for most of the under-represented languages, as shown in Table~\ref{table:personalization}. On the other hand, either the local or the interpolated models are always better than the global one. In other words, the global model alone cannot fairly model data distributions when they are different, and a certain degree of personalization is necessary.
However, a global model enhanced with additional information from other languages can be better when the training examples are overwhelmingly abundant for a particular language. It is shown, for example, that neither the personalized nor the interpolated model can improve over the global model in the case of English, as in Table~\ref{table:personalization}.

It is an open question why in some cases the interpolated model out-performed the personalized version. It maybe the case that our scheme for choosing the interpolation parameter is sub-optimal, or perhaps in some cases the personalized model is over-trained and interpolation provides some regularization with the more general model.  

Finally, our personalization approach assumes a simple handoff from global model training to continued training with local data.  One can imagine other schemes that leverage data from ``nearby'' silos, such as employing Spanish to support a Portuguese model, as the system refines the model towards its target language. This is an interesting avenue for future work.

\subsection{Future Work}
\label{subsec:future}

Federated learning should be considered an indispensable tool in supporting privacy-sensitive data applications, where the training data is distributed and  direct access is severely constrained. The Healthcare setting and scenarios where personalization is needed are among the most prominent FL applications.  HIPAA and other legal frameworks can severely constrain the sharing of private information. and as such, healthcare is one of those industries that can benefit the most from FL.

Although our proposed work addresses some of the technical challenges discussed for a real-life deployment, there is a number of open questions that still need addressing. Among these questions are the computing and bandwidth resources for local training and communication. Further, some participating nodes may be significantly slower, i.e. these nodes are called ``stragglers'',  and solutions based on asynchronous updates are required. The extreme heterogeneity present in the participating clients in the form of data distributions, available training examples have severe impact on the model fairness. Finally, data-related challenges such the lack of good-quality labels can affect the overall performance. Concluding, the number of practical concerns that arises, especially related to quality control and smooth operation, requires additional innovation-- despite the fact that the presented system is production-ready. As part of our future work, our plan is to work and robustify our solution in all of the above technical challenges.

\bibliography{refs.bib}

\begin{thebibliography}{10}
\providecommand{\url}[1]{#1}
\csname url@samestyle\endcsname
\providecommand{\newblock}{\relax}
\providecommand{\bibinfo}[2]{#2}
\providecommand{\BIBentrySTDinterwordspacing}{\spaceskip=0pt\relax}
\providecommand{\BIBentryALTinterwordstretchfactor}{4}
\providecommand{\BIBentryALTinterwordspacing}{\spaceskip=\fontdimen2\font plus
\BIBentryALTinterwordstretchfactor\fontdimen3\font minus
  \fontdimen4\font\relax}
\providecommand{\BIBforeignlanguage}[2]{{%
\expandafter\ifx\csname l@#1\endcsname\relax
\typeout{** WARNING: IEEEtran.bst: No hyphenation pattern has been}%
\typeout{** loaded for the language `#1'. Using the pattern for}%
\typeout{** the default language instead.}%
\else
\language=\csname l@#1\endcsname
\fi
#2}}
\providecommand{\BIBdecl}{\relax}
\BIBdecl

\bibitem{McMahan+16}
H.~B. McMahan, E.~Moore, D.~Ramage, S.~Hampson, and B.~A.~y. Arcas,
  ``Communication-efficient learning of deep networks from decentralized
  data.''

\bibitem{Kairouz+19}
P.~Kairouz and et~al, ``Advances and open problems in federated learning,''
  2019.

\bibitem{HLZ20}
Z.~Huang, F.~Liu, and Y.~Zou, ``Federated learning for spoken language
  understanding,'' in \emph{Proc. of ACL'20}.\hskip 1em plus 0.5em minus
  0.4em\relax ACL, December 2020, pp. 3467--3478.

\bibitem{Wang+22}
H.~Wang, H.~Zhao, Y.~Wang, T.~Yu, J.~Gu, and J.~Gao, ``{FedKC}: Federated
  knowledge composition for multilingual natural language understanding,'' in
  \emph{Proc. of WWW'22}.\hskip 1em plus 0.5em minus 0.4em\relax ACM, April
  2022, p. 1839–1850.

\bibitem{BRSK20}
B.~G. Hb, U.~Reshma, S.~Kp, and M.~Kumar, \emph{MedNLU: Natural Language
  Understander for Medical Texts}, 01 2020, pp. 3--21.

\bibitem{BTRN19}
N.~Bhirud, S.~Tatale, S.~Randive, and S.~Nahar, ``A literature review on
  chatbots in healthcare domain,'' \emph{Int. J. of Scientific \& Technology
  Research}, vol.~8, July 2019.

\bibitem{Liu+21}
M.~Liu, S.~Ho, M.~Wang, L.~Gao, Y.~Jin, and H.~Zhang, ``Federated learning
  meets natural language processing: A survey,'' 2021.

\bibitem{Lin+21}
B.~Y. Lin, C.~He, Z.~Zeng, H.~Wang, Y.~Huang, M.~Soltanolkotabi, X.~Ren, and
  S.~Avestimehr, ``Fednlp: A research platform for federated learning in
  natural language processing,'' 2021.

\bibitem{Yang+18}
T.~Yang, G.~Andrew, H.~Eichner, H.~Sun, W.~Li, N.~Kong, D.~Ramage, , and
  F.~Beaufays, ``Applied federated learning: Improving google keyboard query
  suggestions,'' 2018.

\bibitem{Hard+18}
A.~Hard, K.~Rao, R.~Mathews, S.~Ramaswamy, F.~Beaufays, S.~Augenstein,
  H.~Eichner, C.~Kiddon, and D.~Ramage, ``Federated learning for mobile
  keyboard prediction,'' 2018.

\bibitem{StSi20}
J.~Stremmel and A.~Singh, ``Pretraining federated text models for next word
  prediction,'' 2020.

\bibitem{LHYWZ20}
X.~Li, K.~Huang, W.~Yang, S.~Wang, and Z.~Zhang, ``On the convergence of fedavg
  on non-iid data,'' in \emph{8th Inter. Conf. on Learning Representations,
  {ICLR} 2020}.\hskip 1em plus 0.5em minus 0.4em\relax OpenReview.net, 2020.

\bibitem{Bommasani+22}
R.~Bommasani and et~al, ``On the opportunities and risks of foundation
  models,'' 2022.

\bibitem{DCLT19}
J.~Devlin, M.-W. Chang, K.~Lee, and K.~Toutanova, ``Bert: Pre-training of deep
  bidirectional transformers for language understanding,'' in \emph{Proc. of
  NAACL-HLT'19}.\hskip 1em plus 0.5em minus 0.4em\relax ACM, July 2019.

\bibitem{Brown+20}
T.~B. Brown, B.~Mann, N.~Ryder, and M.~S. et~al., ``Language models are
  few-shot learners,'' 2020.

\bibitem{LiMi20}
D.~Liu and T.~Miller, ``Federated pretraining and fine-tuning of bert using
  clinical notes from multiple silos,'' 2020.

\bibitem{DKMM20}
Y.~Deng, M.~M. Kamani, and M.~Mahdavi, ``Adaptive personalized federated
  learning,'' 2020.

\bibitem{vaswani2017attention}
A.~Vaswani, N.~Shazeer, N.~Parmar, J.~Uszkoreit, L.~Jones, A.~N. Gomez,
  {\L}.~Kaiser, and I.~Polosukhin, ``Attention is all you need,''
  \emph{Advances in neural information processing systems}, vol.~30, 2017.

\bibitem{devlin2018bert}
J.~Devlin, M.-W. Chang, K.~Lee, and K.~Toutanova, ``Bert: Pre-training of deep
  bidirectional transformers for language understanding,'' \emph{arXiv preprint
  arXiv:1810.04805}, 2018.

\bibitem{wang2018glue}
A.~Wang, A.~Singh, J.~Michael, F.~Hill, O.~Levy, and S.~R. Bowman, ``Glue: A
  multi-task benchmark and analysis platform for natural language
  understanding,'' \emph{arXiv preprint arXiv:1804.07461}, 2018.

\bibitem{wang2019superglue}
A.~Wang, Y.~Pruksachatkun, N.~Nangia, A.~Singh, J.~Michael, F.~Hill, O.~Levy,
  and S.~Bowman, ``Superglue: A stickier benchmark for general-purpose language
  understanding systems,'' \emph{Advances in neural information processing
  systems}, vol.~32, 2019.

\bibitem{gururangan2020don}
S.~Gururangan, A.~Marasovi{\'c}, S.~Swayamdipta, K.~Lo, I.~Beltagy, D.~Downey,
  and N.~A. Smith, ``Don't stop pretraining: adapt language models to domains
  and tasks,'' \emph{arXiv preprint arXiv:2004.10964}, 2020.

\bibitem{liang2020xglue}
Y.~Liang, N.~Duan, Y.~Gong, N.~Wu, F.~Guo, W.~Qi, M.~Gong, L.~Shou, D.~Jiang,
  G.~Cao \emph{et~al.}, ``Xglue: A new benchmark dataset for cross-lingual
  pre-training, understanding and generation,'' \emph{arXiv preprint
  arXiv:2004.01401}, 2020.

\bibitem{conneau2019unsupervised}
A.~Conneau, K.~Khandelwal, N.~Goyal, V.~Chaudhary, G.~Wenzek, F.~Guzm{\'a}n,
  E.~Grave, M.~Ott, L.~Zettlemoyer, and V.~Stoyanov, ``Unsupervised
  cross-lingual representation learning at scale,'' \emph{arXiv preprint
  arXiv:1912.02116}, 2019.

\bibitem{jiang2021xlm}
X.~Jiang, Y.~Liang, W.~Chen, and N.~Duan, ``Xlm-k: Improving cross-lingual
  language model pre-training with multilingual knowledge,'' in
  \emph{Proceedings of the AAAI Conference on Artificial Intelligence},
  vol.~36, no.~10, 2022, pp. 10\,840--10\,848.

\bibitem{beltagy2019scibert}
I.~Beltagy, K.~Lo, and A.~Cohan, ``Scibert: A pretrained language model for
  scientific text,'' \emph{arXiv preprint arXiv:1903.10676}, 2019.

\bibitem{lee2020biobert}
J.~Lee, W.~Yoon, S.~Kim, D.~Kim, S.~Kim, C.~H. So, and J.~Kang, ``Biobert: a
  pre-trained biomedical language representation model for biomedical text
  mining,'' \emph{Bioinformatics}, vol.~36, no.~4, pp. 1234--1240, 2020.

\bibitem{pubmedbert}
Y.~Gu, R.~Tinn, H.~Cheng, M.~Lucas, N.~Usuyama, X.~Liu, T.~Naumann, J.~Gao, and
  H.~Poon, ``Domain-specific language model pretraining for biomedical natural
  language processing,'' 2020.

\bibitem{bird2009natural}
S.~Bird, E.~Klein, and E.~Loper, \emph{Natural language processing with Python:
  analyzing text with the natural language toolkit}.\hskip 1em plus 0.5em minus
  0.4em\relax " O'Reilly Media, Inc.", 2009.

\bibitem{magnini2020e3c}
B.~Magnini, B.~Altuna, A.~Lavelli, M.~Speranza, and R.~Zanoli, ``The e3c
  project: Collection and annotation of a multilingual corpus of clinical
  cases,'' in \emph{CLiC-it}, 2020.

\bibitem{gotzsche2011opening}
P.~C. G{\o}tzsche and A.~W. J{\o}rgensen, ``Opening up data at the european
  medicines agency,'' \emph{Bmj}, vol. 342, 2011.

\bibitem{wikipedia2004wikipedia}
Wikipedia, \emph{Wikipedia}.\hskip 1em plus 0.5em minus 0.4em\relax PediaPress,
  2004.

\bibitem{roberts2001pubmed}
R.~J. Roberts, ``Pubmed central: The genbank of the published literature,'' pp.
  381--382, 2001.

\bibitem{neveol2014quaero}
A.~N{\'e}v{\'e}ol, C.~Grouin, J.~Leixa, S.~Rosset, and P.~Zweigenbaum, ``The
  quaero french medical corpus: A ressource for medical entity recognition and
  normalization,'' in \emph{In proc biotextm, reykjavik}.\hskip 1em plus 0.5em
  minus 0.4em\relax Citeseer, 2014.

\bibitem{packer2009scielo}
A.~L. Packer, ``The scielo open access: a gold way from the south.''
  \emph{Canadian Journal of Higher Education}, vol.~39, no.~3, pp. 111--126,
  2009.

\bibitem{miranda2020codiesp}
A.~Miranda-Escalada and A.~Gonzalez-Agirre, ``Codiesp: Clinical case coding in
  spanish shared task (ehealth clef 2020),'' in \emph{eHealth CLEF 2020}, 2020.

\bibitem{zarin2011clinicaltrials}
D.~A. Zarin, T.~Tse, R.~J. Williams, R.~M. Califf, and N.~C. Ide, ``The
  clinicaltrials. gov results database—update and key issues,'' \emph{New
  England Journal of Medicine}, vol. 364, no.~9, pp. 852--860, 2011.

\bibitem{johnson2016mimic}
A.~E. Johnson, T.~J. Pollard, L.~Shen, L.-w.~H. Lehman, M.~Feng, M.~Ghassemi,
  B.~Moody, P.~Szolovits, L.~Anthony~Celi, and R.~G. Mark, ``Mimic-iii, a
  freely accessible critical care database,'' \emph{Scientific data}, vol.~3,
  no.~1, pp. 1--9, 2016.

\bibitem{johnson2019mimic}
A.~Johnson, T.~Pollard, R.~Mark, S.~Berkowitz, and S.~Horng, ``Mimic-cxr
  database,'' \emph{PhysioNet10}, vol. 13026, p. C2JT1Q, 2019.

\bibitem{uzuner2006i2b2}
O.~Uzuner, P.~Szolovits, and I.~Kohane, ``i2b2 workshop on natural language
  processing challenges for clinical records,'' in \emph{Proceedings of the
  Fall Symposium of the American Medical Informatics Association}.\hskip 1em
  plus 0.5em minus 0.4em\relax Citeseer, 2006.

\bibitem{tiedemann2012parallel}
J.~Tiedemann, ``Parallel data, tools and interfaces in opus.'' in \emph{Lrec},
  vol. 2012, 2012, pp. 2214--2218.

\bibitem{ima}
I.~M. Association, ``Israeli medical association,''
  \url{https://www.ima.org.il/}.

\bibitem{mytrail}
I.~C. Trail, ``Israeli clinical trail,''
  \url{https://www.gov.il/he/departments/general/clinical-trials-website}.

\bibitem{Dimitriadis+21}
D.~Dimitriadis, K.~Kumatani, R.~Gmyr, Y.~Gaur, and S.~E. Eskimez, ``Dynamic
  gradient aggregation for federated domain adaptation,'' 2021.

\bibitem{CFGN96}
R.~Canetti, U.~Feige, O.~Goldreich, and M.~Naor, ``Adaptively secure
  multi-party computation,'' in \emph{In proc. of Symp. on Theory of Computing
  (STOC ’96)}, 1996.

\bibitem{BCDR}
\BIBentryALTinterwordspacing
J.~Li, Y.~Sun, R.~J. Johnson, D.~Sciaky, C.~Wei, R.~Leaman, A.~P. Davis, C.~J.
  Mattingly, T.~C. Wiegers, and Z.~Lu, ``Biocreative {V} {CDR} task corpus: a
  resource for chemical disease relation extraction,'' \emph{Database J. Biol.
  Databases Curation}, vol. 2016, 2016. [Online]. Available:
  \url{https://doi.org/10.1093/database/baw068}
\BIBentrySTDinterwordspacing

\bibitem{dougan2014ncbi}
R.~I. Do{\u{g}}an, R.~Leaman, and Z.~Lu, ``Ncbi disease corpus: a resource for
  disease name recognition and concept normalization,'' \emph{Journal of
  biomedical informatics}, vol.~47, pp. 1--10, 2014.

\bibitem{patrick2010high}
J.~Patrick and M.~Li, ``High accuracy information extraction of medication
  information from clinical notes: 2009 i2b2 medication extraction challenge,''
  \emph{Journal of the American Medical Informatics Association}, vol.~17,
  no.~5, pp. 524--527, 2010.

\bibitem{segura2013semeval}
I.~Segura-Bedmar, P.~Mart{\'\i}nez~Fern{\'a}ndez, and M.~Herrero~Zazo,
  ``Semeval-2013 task 9: Extraction of drug-drug interactions from biomedical
  texts (ddiextraction 2013).''\hskip 1em plus 0.5em minus 0.4em\relax
  Association for Computational Linguistics, 2013.

\bibitem{dimitriadis2022flute}
D.~Dimitriadis, M.~H. Garcia, D.~M. Diaz, A.~Manoel, and R.~Sim, ``Flute: A
  scalable, extensible framework for high-performance federated learning
  simulations,'' \emph{arXiv preprint arXiv:2203.13789}, 2022.

\bibitem{SMS20}
F.~Sattler, K.-R. Müller, and W.~Samek, ``Clustered federated learning:
  Model-agnostic distributed multitask optimization under privacy
  constraints,'' \emph{IEEE Trans. on Neural Networks and Learning Systems},
  pp. 1--13, 08 2020.

\bibitem{TYCY22}
A.~Z. Tan, H.~Yu, L.~Cui, and Q.~Yang, ``Towards personalized federated
  learning,'' \emph{IEEE Trans. on Neural Networks and Learning Systems}, pp.
  1--17, 03 2022.

\end{thebibliography}

\end{document}